\def\year{2022}\relax
\documentclass[letterpaper]{article} 
\usepackage{aaai22}  
\usepackage{times}  
\usepackage{helvet}  
\usepackage{courier}  
\usepackage[hyphens]{url}  
\usepackage{graphicx} 
\usepackage{caption} 
\DeclareCaptionStyle{ruled}{labelfont=normalfont,labelsep=colon,strut=off} 
\usepackage{algorithm}

\usepackage{xcolor}         
\usepackage[utf8]{inputenc} 
\usepackage[T1]{fontenc}    
\usepackage{hyperref}       
\usepackage{url}            
\usepackage{booktabs}       
\usepackage{tablefootnote}
\usepackage{amsfonts}       
\usepackage{nicefrac}       
\usepackage{microtype}      

\usepackage{hyperref}       
\usepackage{url}            
\usepackage{booktabs}       
\usepackage{amsfonts}       
\usepackage{nicefrac}       
\usepackage{microtype}      
\usepackage{lipsum}

\usepackage{mathtools} 
\usepackage{amsmath}
\usepackage{algorithm}
\usepackage{algpseudocode}
\usepackage{amssymb}
\usepackage{graphicx}
\usepackage{amsmath}
\usepackage{enumitem}

\usepackage{bbold}
\usepackage{subfig}

\usepackage{multirow}
\usepackage{dsfont}

\usepackage{caption}

\usepackage{booktabs,ragged2e}

\usepackage[flushleft]{threeparttable}

\usepackage[export]{adjustbox}

\makeatletter
\newcounter{phase}[algorithm]
\newlength{\phaserulewidth}
\newcommand{\setphaserulewidth}{\setlength{\phaserulewidth}}

\makeatother
\setphaserulewidth{.7pt}

\algdef{SE}[SUBALG]{Indent}{EndIndent}{}{\algorithmicend\ }%
\algtext*{Indent}
\algtext*{EndIndent}

\usepackage{amsthm}
\usepackage{wrapfig}

\newlength{\tempheight}
\newlength{\tempwidth}

\newcommand{\rowname}[1]
{\rotatebox{90}{\makebox[\tempheight][c]{\textbf{#1}}}}

\newcommand{\columnname}[1]
{\makebox[\tempwidth][c]{\textbf{#1}}}

\usepackage{array}
\makeatletter
\g@addto@macro{\endtabular}{\gdef\rowfonttype{}}
\makeatother
\newcommand{\rowfonttype}{}
\newcolumntype{L}{>{\rowfonttype\strut}l}
\newcolumntype{C}{>{\rowfonttype\strut}c}
\newcolumntype{R}{>{\rowfonttype\strut}r}

\usepackage{newfloat}
\usepackage{listings}
\floatstyle{ruled}
\newfloat{listing}{tb}{lst}{}
\floatname{listing}{Listing}
\title{Information Theoretic Structured Generative Modeling}
\author{Bo~Hu\textsuperscript{\rm 1}, Shujian Yu\textsuperscript{\rm 2, \rm 3}, Jos{\'e}~C.~Pr{\'\i}ncipe\textsuperscript{1} }
\affiliations{
    \textsuperscript{\rm 1}University of Florida\\
    Gainesville, FL 32611, USA\\
    hubo@ufl.edu; principe@cnel.ufl.edu \\[1mm]
    \textsuperscript{\rm 2}UiT - The Arctic University of Norway, 9037 Troms{\o}, Norway\\
    \textsuperscript{\rm 3}Xi'an Jiaotong University, Xi'an $710049$, Shaanxi, China\\
    {yusj9011@gmail.com}
}

\usepackage{bibentry}

\begin{document}

\maketitle

\begin{abstract}
R\'enyi's information provides a theoretical foundation for tractable and data-efficient non-parametric density estimation, based on pair-wise evaluations in a reproducing kernel Hilbert space (RKHS). This paper extends this framework to parametric probabilistic modeling, motivated by the fact that R\'enyi's information can be estimated in closed-form for Gaussian mixtures. Based on this special connection, a novel generative model framework called the structured generative model (SGM) is proposed that makes straightforward optimization possible, because costs are scale-invariant, avoiding high gradient variance while imposing less restrictions on absolute continuity, which is a huge advantage in parametric information theoretic optimization. The implementation employs a single neural network driven by an orthonormal input appended to a single white noise source adapted to learn an infinite Gaussian mixture model (IMoG), which provides an empirically tractable model distribution in low dimensions. To train SGM, we provide three novel variational cost functions, based on R\'enyi's second-order entropy and divergence, to implement minimization of cross-entropy, minimization of variational representations of $f$-divergence, and maximization of the evidence lower bound (conditional probability). We test the framework for estimation of mutual information and compare the results with the mutual information neural estimation (MINE), for density estimation, for conditional probability estimation in Markov models as well as for training adversarial networks. Our preliminary results show that SGM significantly improves MINE estimation in terms of data efficiency and variance, conventional and variational Gaussian mixture models, as well as the performance of generative adversarial networks.
\end{abstract}

\section{Introduction}
The traditional information descriptors in probability spaces are Shannon's entropy, Mutual Information, and the Kullback–Leibler divergence (relative entropy). Since the probability density function of the data in machine learning is often unknown, applying these descriptors requires well-behaved estimators or approximations. The objective of information theoretic learning (ITL) is to provide a clear and theoretically supported view of probabilistic modeling algorithms. There are basically two different avenues in the literature: non-parametric estimators and parametric estimators. R\'enyi's second-order entropy and divergence have been broadly used for designing non-parametric estimators~\cite{principebook, luis} following the principle of sample-based pair-wise evaluations in reproducing kernel Hilbert space (RKHS). The main difficulty of this non-parametric framework is its dependence on a free parameter that yields a bias. If the bias can be tolerated (e.g. optimization of ITL cost functions), its appeal is interpretability, data efficiency, and good performance \cite{shujianinformation}.

The other avenue optimizes a universal parametric model to estimate the ITL descriptors, which usually follows three types of principles: minimizing cross-entropy~\cite{bengioentropyminimum, bishopmixture}, minimizing variational representations of $f$-divergence~\cite{jordan1, jordan2, gan}, and maximizing the evidence lower bound (ELBO)~\cite{blei, imog, li2016renyi}. For example, minimization of cross-entropy, which is simple to implement from the data distributions, yields a model that is an upper bound for the data entropy ~\cite{bengioentropyminimum, bishopmixture}. Solving variational representations of $f$-divergence based on a function of the probability ratio $p(x)/q(x)$ quantifies the divergence between two distributions as an upper bound of the model mapping ~\cite{jordan1, jordan2}. Recently, generative adversarial networks (GANs)~\cite{gan} take this approach with one network approximating $f$-divergence and another network producing an intractable model distribution to minimize the approximated $f$-divergence~\cite{wgan,fgan}. In variational inference, if direct optimization over cross-entropy is difficult, ELBO introduces a latent variable with conditional probabilities to optimizes the parameters of an underlying model that has a simpler structure~\cite{blei}.

However, experience has shown that optimizing all these costs is a challenging task. The main reason is that Shannon's cross-entropy or Kullback–Leibler divergence require absolute continuity of two Lebesgue measures $Q\gg P$, which is highly restrictive during adaptation of any parameterized model that yields the distribution. Without this condition the bounds do not hold. Secondly, in order to avoid estimating the nonlinear $\log q(x)$, the asymptotic equipartition property (AEP) is normally utilized~\cite{Liamestimate}, which is not data efficient. Thirdly, the ultimate task is to produce a valid probability distribution, which requires constrained optimization methodologies. All three aspects also create high variance in the gradient and forbid the application of simple optimization techniques. The field conventionally uses expectation maximization (EM) ~\cite{bishopmixture}, stochastic approximation~\cite{jordanmeanfield}, or what is also called ``reparameterization tricks''~\cite{kingma2013auto}. 

The main contribution of this paper is to propose a unified new set of information theoretic cost functions based on the theory of R\'enyi's entropy and divergence~\cite{Renyi} that improves these three issues.  R\'enyi's entropy and the corresponding divergence are related to the special case of $f$-divergence when the convex function $f$ is taken to be a polynomial. We derive three variational forms for R\'enyi's second-order entropy, divergence, and conditional entropy that corresponds to the three mentioned principles. The important property of R\'enyi's information as opposed to Shannon is that the log is outside of the integral, which simplifies the estimation~\cite{principebook} and provides a closed-form solution for mixture of Gaussian~\cite{kampa2011closed}, which dramatically simplifies the computation for the versatile class of generalized Gaussian mixture models. The second contribution is to propose a new generative model framework called the structured generative model (SGM) for modeling distributions or conditional distributions by approximating local sample densities by their mean and variance, assuming a versatile generalized Gaussian mixture model. SGM uses a single neural network to produce an infinite mixture of Gaussian (IMoG) that can be trained efficiently and robustly by the newly proposed cost through gradient methods. Compared to conventional methods, it can produce an empirically tractable probability distribution and hugely improves performance with much smaller bias and less variance for the class of IMoG problems. To further demonstrate the superiority of our cost function, we compare with MINE estimators, we show its versatility by estimating conditional probabilities, and we also show an example of the newly proposed variational divergence form to train a GAN. Under the proposed framework, the discriminator network will produce a probability ratio, which can have broad applications such as out-of-distribution (OOD) detection~\cite{hendrycks2016baseline}.

\section{Background}\label{section2}

We start with an introduction of cross-entropy, $f$-divergence, and a short discussion about Gaussian mixture models (GMM) and generative adversarial networks (GAN) from the perspective of ITL. Then we introduce R\'enyi's entropy and divergence. Throughout the paper, we assume that the density functions exist and are Lebesgue measurable. We use the definition of differential entropy for the continuous random variables.\\

\noindent \textbf{Cross-entropy and GMM}: Let a density function $q(x)$ be given. Taking the expectation of $-\log q(x)$ over $p(x)$ yields the cross-entropy
\begin{equation}
\begin{aligned}
\text{CE}(p, q) &= \int_{\mathcal{X}}p(x)\log p(x) + \text{KL}(p\|q) \\&\geq -\int_{\mathcal{X}}p(x)\log p(x) = H(p).
\label{CE}
\end{aligned}
\end{equation}
Thus, minimizing $\text{CE}(p, q)$ yields $H(p)$, and the tightness of the bounds depends directly on $\text{KL}(p\|q)$. It is well-known that Gaussian mixture models maximizes the log likelihood of data, but it can also be formulated as minimizing the cross-entropy. Let the model density be $g_\theta(x) = \Sigma_{i=1}^N w_i\mathcal{N}(x-m_i, A_i)$. By~\eqref{CE}, the cross-entropy satisfies $\text{CE}(p, g_\theta) = -\int_\mathcal{X} p(x) \log \Sigma_{i=1}^N w_i\mathcal{N}(x-m_i, A_i) d\mu \geq H(p)$. We write $g_\theta(x, z = i) = w_i\mathcal{N}(x-m_i, A_i)$. Since ${\partial \text{CE}(p, g_\theta)}/{\partial \theta} = 0$ does not have a closed-form solution, expectation maximization (EM) and variational inference optimize an upper bound of $\text{CE}(p, g_\theta)$:
\begin{equation}
\begin{aligned}
\text{CE}(p, g_\theta) \leq \text{CE}\Big(p(x)\frac{g_\theta(x, z)}{g_\theta(x)}, g_\theta(x, z)\Big),
\end{aligned}
\end{equation}
Ideally, the upper bound is tight when $p(x){g_\theta(x, z)}/{ g_\theta(x)} = g_\theta(x, z)$, i.e., $g_\theta(x) = p(x)$, which yields the same solution as minimizing $\text{CE}(p, g_\theta)$. EM iteratively updates $\theta^{new} = \max_{\theta}\text{CE}\big(p(x){g_{\theta'}(x, z)}/{g_{\theta'}(x)}, g_\theta(x, z)\big)$. Gradient methods can also be used~\cite{bishopmixture, jin2016local}. However, the condition for this upper bound to hold is very restrictive. Suppose there exists $x$ such that $p(x)>0$ and $g_\theta(x) = 0$, it is easy to show that $\text{CE}(p, g_\theta)\rightarrow \infty$. On the other hand, the responsibility $g_\theta(x, z)/g_\theta(x)$ is taken to be finite such that the RHS has finite values. In this case, this bound no longer exists. \\

\noindent \textbf{$f$-divergence and GAN}: Another important branch of costs is the $f$-divergence. Given a convex function $f$ with $f(1)=0$, $f$-divergence has the form $D_f(p \| q) = \int_\mathcal{X} q(x)f(p(x)/q(x)) d\mu$ that evaluates a ``distance '' between two distributions. A variational form of $D_f(p \| q)$ takes the convex conjugate of $f$~\cite{hiriart2004fundamentals}, which becomes~\cite{fgan}:
\begin{equation}
D_f(p \| q) \geq \sup_{T\in\mathcal{T}} (\mathbb{E}_p [T(x)] -   \mathbb{E}_q [f^*(T(x))]),
\label{fdivergence}
\end{equation}
where $\mathcal{T}$ is an arbitrary class of functions $T:\mathcal{X}\rightarrow \mathbb{R}$. The bound is tight when $T_0(x) = f'({p(x)}/{q(x)})$. Although $T_0$ is written in the form of ${p(x)}/{q(x)}$, the absolute continuity condition also depends on the choice of $f$.

Any universal parametric model such as kernel functions~\cite{jordan1, jordan2} or neural networks~\cite{mine} can be used to approximate $T$ by maximizing the lower bound. GAN uses a discriminator network to approximate $T$, and a generator network to produce an intractable model distribution $g_\theta$ that minimizes the approximated $f$-divergence~\cite{wgan, fgan}. The most obvious downside is that $g_\theta$ is intractable. Secondly, although $T_\theta$ should approach $T_0(x)$ related to the probability ratio, there is no guarantee in GAN.\\

\noindent \textbf{R\'enyi's entropy and divergence}: We also mention R\'enyi's entropy and divergence of order $\alpha$
\begin{equation}
\begin{gathered}
H_\alpha(p) = \frac{1}{1-\alpha}\log \int_\mathcal{X} {p^\alpha(x)} d\mu, \\
D_\alpha(p\|q) = \frac{1}{1-\alpha}\log \int_\mathcal{X} {p^\alpha(x)q^{1-\alpha}(x)} d\mu.
\label{ren}
\end{gathered}
\end{equation}
R\'enyi's divergence is related to the special case of \eqref{fdivergence} when $f$ is chosen to be a polynomial. The existence of $D_\alpha(p\|q)$ requires $Q\gg P$ for all $\alpha \geq 1$. When $\alpha$ is 2, R\'enyi's second-order divergence matches the $\chi^2$-divergence. R\'enyi's second-order entropy and divergence are the foundations of non-parametric density estimators based on sample-based pair-wise evaluations and the theory of RKHS~\cite{principebook}. This paper further expands this idea.\\

\noindent \textbf{Properties of MoG}: Finally, we mention of pair-wise properties of mixture of Gaussian (MoG) that we utilize for the design. Given any two Gaussian density functions, it can be shown that $\int_\mathcal{X} \mathcal{N}(x-m_i, A_{i}) \mathcal{N}(x-m_j, A_{j}) d\mu = \mathcal{N}(m_i-m_j, A_i + A_j)$. Now suppose we have a MoG with a density function $T_\varphi(x) = \Sigma_{i=1}^N w_i \mathcal{N}(x-m_i, A_{i})$. It follows that $T_\varphi(x)$ satisfies
\begin{equation}
\begin{aligned}
\int_\mathcal{X} T_\varphi^2(x) d\mu = \Sigma_{i=1}^N \Sigma_{j=1}^N w_i w_j \mathcal{N}(m_i-m_j, A_i + A_j),
\end{aligned}
\label{discrete}
\end{equation}
which is fully determined by $\varphi$ with a closed-form solution based on pair-wise relations~\cite{kampa2011closed}.

We can further show that the same conclusion holds for an infinite mixture of Gaussian (IMoG) in a Bayesian setting, where $\varphi$ is determined by an arbitrary distribution $P_\varphi$. Thus the model density has the form $T_\varphi(x) = \mathbb{E}_{\varphi \sim P_\varphi}\big[ \mathbf{w} \mathcal{N}(x-\mathbf{m}, \mathbf{A}) \big]$. Similarly it can be shown that
\begin{equation}
\resizebox{1\linewidth}{!}{
$\begin{aligned}
\int_\mathcal{X} T_\varphi^2(x) d\mu = \mathbb{E}_{\varphi_1\sim P_{\varphi}, \varphi_2 \sim P_{\varphi}}\big[ \mathbf{w}_1 \mathbf{w}_2 \mathcal{N}(\mathbf{m}_1-\mathbf{m}_2, \mathbf{A}_1+ \mathbf{A}_2) \big].
\label{qudratic_form}
\end{aligned}$}
\end{equation}
In the following sections, we show how we use a neural network to produce the model density $T_\varphi(x)$. While imposing a MoG can introduce a bias~\cite{bishopmixture}, the advantage of controlling the variance with a model is very appealing and surpasses the former.

\section{Proposed Cost Functions}
In this section, we introduce two new variational cost functions $J_{p}(T)$ and $J_{p, q}(T)$, which accepts upper bound related to R{\'e}nyi's second-order entropy $H_2(P)$ and second-order divergence $D_{2}(P \| Q)$. \\

\noindent \textbf{Variational R\'enyi's second-order entropy:} Similar to the cross-entropy, we introduce a functional $J_{p}(T)$ with the form
\begin{equation}
J_{p}(T) = \frac{\mathbb{E}_p[T(x)]}{({\int_\mathcal{X} T^2(x) d\mu})^{\frac{1}{2}}},
\label{cost1}
\end{equation}
over the set $\mathcal{T}$ that contains all non-negative measurable functions with $\int_\mathcal{X} T^2(x) d\mu>0$. For simplicity, we write the inner product $\int_\mathcal{X} p(x)T(x) d\mu := {\langle p, T\rangle}$, the norm $\int_{\mathcal{X}}p^2(x)d\mu = {\langle p, p \rangle}$, and $\int_{\mathcal{X}}T^2(x)d\mu = { {\langle T, T\rangle} }$. By Cauchy-Schwarz inequality, we have $\langle p, T\rangle \leq \sqrt{\langle p, p\rangle \langle T, T\rangle}$. It follows that $J_{p}(T)= {\langle p, T\rangle}/{ \sqrt{\langle T, T\rangle} } \leq \sqrt{\langle p, p \rangle}$. Therefore we obtained
\begin{equation}
\sup_{T \in \mathcal{T}} J_p(T) = \Big({\int_{\mathcal{X}}p^2(x)d\mu}\Big)^{\frac{1}{2}}.
\label{upperbound}
\end{equation}
There are two important conclusions. First, the upper bound only depends the data. Second, the solution is $T_0(x) = \beta p(x)$ with $\beta$ taken to be any arbitrary positive scalar. The function $T$ can be approximated by any parametric model.\\

\noindent \textbf{R\'enyi's second-order divergence:} Following a similar idea, we introduce a functional $J_{p, q}(T)$ for R\'enyi's second-order divergence, written as 
\begin{equation}
J_{p, q}(T) =  \frac{\mathbb{E}_p[T(x)]}{({\mathbb{E}_q[T^2(x)]})^{\frac{1}{2}}}.
\label{cost2}
\end{equation}
Similarly, by appplying Cauchy-Schwartz inequality,
\begin{equation}
J_{p, q}(T) = \frac{\langle \frac{p(x)}{\sqrt{q(x)}}, T {\sqrt{q(x)}}\rangle}{{\langle T {\sqrt{q(x)}}, T {\sqrt{q(x)}}\rangle}^{\frac{1}{2}}} \leq {\langle  {\frac{p(x)}{\sqrt{q(x)}}},  {\frac{p(x)}{\sqrt{q(x)}}}\rangle}^{\frac{1}{2}}.
\end{equation}
Therefore we proved that there is an upper bound for $J_{p, q}(T)$
\begin{equation}
\sup_{T\in \mathcal{T}} J_{p, q}(T) = \Big({\int_{\mathcal{X}} \frac{{p^2(x)}}{q(x)}d\mu}\Big)^{\frac{1}{2}}.
\end{equation}
The supremum is attained as $T_0(x) = \beta {p(x)}/{q(x)}$, which is a scaled value of the probability ratio. Combining with \eqref{ren}, it can be shown that the upper bounds of $J_{p}(T)$ and $J_{p, q}(T)$ are related to R\'enyi's second-order entropy and divergence as $H_2(P) = -2 \log \sup_{T\in \mathcal{T}}J_{p}(T)$ and $D_{2}(P\|Q) = -2 \log \sup_{T\in \mathcal{T}}J_{p, q}(T)$.

\section{Proposed Algorithms}\label{4.2}
Now we show how  $J_p(T)$ and $J_{p, q}(T)$ can be employed for model training. First, optimizing $J_{p, q}(T)$ shares the same procedure as optimizing any variational form of $f$-divergence~\cite{jordan1, jordan2, mine}. It can also train a generative adversarial network with the discriminator network producing the probability ratio. Optimizing $J_p(T)$ is challenging because the normalizing term $\int_\mathcal{X} T^2(x) d\mu$. One unwise choice is to impose a uniform distribution such that $\int_\mathcal{X} T^2(x) d\mu = \mathbb{E}_u[T^2(x)]/Z$. This is not practical since the support of the data distribution may be unknown. Therefore, we use the properties of MoG introduced in the Background section that if $T(x)$ is a mixture of Gaussian (MoG) or an infinite mixture of Gaussian (IMoG), the term $\int_\mathcal{X} T^2(x) d\mu$ will have a closed-form solution~\eqref{qudratic_form}. In this section, we show how we use a neural network to produce this density function of IMoG. We first propose the following density approximation algorithm based on this property.\\

\noindent \textbf{Density approximation:} Similar to GAN, we impose two types of network inputs $\mathbf{z}$ and $\mathbf{c}$. We first define the discrete orthogonal vectors $\mathbf{z}$. We define a vector $z_i = [z_{i}(1), z_{i}(2), \cdots, z_{i}(N)]^\intercal$, with each element satisfying $z_{i}(j)=
\begin{cases} 
1 & \text{if }i=j\\ 
0 & \text{if }i\neq j 
\end{cases}$. Now we obtained a orthonormal set of $N$ one-hot vectors $\{z_1, z_2, \cdots, z_N\}$. As for $\mathbf{c}$, we assume it is sampled from a uniform distribution $\mathbf{c}\sim U(0, 1)$. We call the combination of the orthonormal set and the uniformly distributed noise the scanning vectors, whose role is to span the sample space to quantify local structures. The scanning vectors form the inputs to the neural network.

The output of the neural net defines $P_\varphi$. With $z_i$ and $\mathbf{c}$ as the input, we denote the output of the neural net as $\{f^{(w)}_\theta(z_i, \mathbf{c}),  f^{(m)}_\theta(z_i, \mathbf{c}), f^{(A)}_\theta(z_i, \mathbf{c}))\}$, which defines the model density function
\begin{equation}
\resizebox{1\linewidth}{!}{
$g_\theta(x) = \mathbb{E}_\mathbf{c} \big[\frac{1}{N}\Sigma_{i=1}^N f^{(w)}_\theta(z_i, \mathbf{c}) \mathcal{N}(x-f^{(m)}_\theta(z_i, \mathbf{c}), f^{(A)}_\theta(z_i, \mathbf{c})) \big].$
\label{infinite_gauss_form}}
\end{equation}
Rewriting $w_i(\mathbf{c}) = f^{(w)}_\theta(z_i, \mathbf{c})$, $
m_i(\mathbf{c}) = f^{(m)}_\theta(z_i, \mathbf{c})$, and $
A_i(\mathbf{c}) = f^{(A)}_\theta(z_i, \mathbf{c})$. With \eqref{qudratic_form}, we have
\begin{equation}
\begin{gathered}
\mathcal{N}_{i, j} = \mathcal{N}\big(m_i(\mathbf{c}_1) - m_j(\mathbf{c}_2), A_i(\mathbf{c}_1) + A_j(\mathbf{c}_2) \big)\\
\int_{\mathcal{X}}g_\theta^2(x) d\mu  =  \mathbb{E}_{\mathbf{c}_1, \mathbf{c}_2} \big[\frac{1}{N^2}\Sigma_{i, j=1}^{N} w_i(\mathbf{c}_1) w_j(\mathbf{c}_2 ) \mathcal{N}_{i, j}  \big],
\end{gathered}
\end{equation}
which is fully defined by $P_\varphi$, irrelevant to the data distribution and avoids an empirical estimation over $\mathcal{X}$. The optimization problem can be written as
\begin{equation}
\text{maximize}_{g_\theta} \;\; J_p(g_\theta) =  \frac{\mathbb{E}_p[g_\theta(x)]}{({\int_\mathcal{X} g^2_\theta(x) d\mu})^{\frac{1}{2}}}.
\label{global}
\end{equation}
The added advantage is that the model can be trained efficiently by gradient ascent. The upper bound of the cost function is given by~\eqref{upperbound} and the bound is tight if $g_\theta(x) = p(x)$. We call this approach the structured generative model (SGM). 

One difficulty for optimizing directly with such a cost is the variance of the gradient, which can be reduced in a procedure introduced in~\cite{9415011}. We define $k_\theta = {\mathbb{E}_p[g_\theta(x)]}$ and $v_\theta = \int_\mathcal{X} g^2_\theta(x) d\mu$, the gradient of $J_p(g_\theta)$ has the form $\frac{\partial J_p(g_\theta)}{\partial \theta} = \Big(\frac{\partial k_\theta}{\partial \theta}/\sqrt{v_\theta}\Big) -  \frac{1}{2}\Big((k_\theta  \frac{\partial v_\theta}{\partial \theta} )/\sqrt{v_\theta^3}\Big)$. Then $k_\theta$ and $v_\theta$ can be estimated adaptively over time. At iteration~$t$, suppose the adaptive filters produce $\Tilde{k}_t$ and $\Tilde{v}_t$. Suppose the empirical estimations based on $\theta_t$ are $\hat{k}_{\theta_t}$ and $\hat{v}_{\theta_t}$. The gradient can be constructed by $\frac{\partial J_p(g_{\theta_t})}{\partial {\theta_t}} \approx \Big(\frac{\partial \hat{k}_{\theta_t}}{\partial {\theta_t}}/\sqrt{\Tilde{v}_t}\Big) -  \frac{1}{2}\Big((\Tilde{k}_t \frac{\partial \hat{v}_{\theta_t}}{\partial {\theta_t}} )/\sqrt{\Tilde{v}_t^3} \Big)$. This trick has been also applies to the divergence cost~\eqref{cost2}. Now we can present the full algorithm of SGM for density estimation in Algorithm~\ref{mainalgorithm}.
\begin{algorithm}
\small
\caption{SGM for Density Estimation}
\begin{algorithmic}[1]
\State Initialize $k_0\leftarrow 0$; $v_0 \leftarrow 0$; $t \leftarrow 0$; Initialize $\theta_0$
\While{$\theta_t$ not converge}
\State $t \leftarrow t+1$
\State Sample $\{c_1, c_2, \cdots, c_{M}\}$ from $U(0, 1)$; Sample $\{x_1, x_2, \cdots, x_{M}\}$ from data distribution
\For{$i = 1, 2 \cdots N$ and {$j = 1, 2 \cdots M$}} 
\State $w_{i, j} = f_{\theta_{t-1}}^{(w)}(z_i, c_j)$; $m_{i, j} = f_{\theta_{t-1}}^{(m)}(z_i, c_j)$; $A_{i, j} = f_{\theta_{t-1}}^{(A)}(z_i, c_j)$
\EndFor
\State $v_t' = \frac{1}{M^2N^2}\Sigma_{i,j = 1}^{N,M} \Sigma_{i',j' = 1}^{N,M}  w_{i,j} w_{i',j'} \mathcal{N}(m_{i,j} - m_{i',j'}, A_{i, j}+A_{i', j'})$
\State $k_t' = \frac{1}{MN^2}\Sigma_{i,j = 1}^{N,M} \Sigma_{j' = 1}^{M} w_{i,j}\mathcal{N}(m_{i,j} - x_{j'}, A_{i,j})$
\State $v_t = \beta_1 \cdot v_{t-1} + (1-\beta_1) \cdot v_t' $
\State $\hat{v}_t = v_t/(1-\beta_1^t)$
\State $k_t = \beta_2 \cdot k_{t-1} + (1-\beta_2) k_t'$
\State $\hat{k}_t = k_t/(1-\beta_2^t)$
\State $\theta_{t} = \theta_{t-1} + r \big(  \frac{\partial k_t'}{\partial {\theta_t}}/\sqrt{\hat{v}_t} -  \frac{1}{2}(\hat{k}_t \frac{\partial v_t'}{{\theta_t}} )/\sqrt{\hat{v}_t^3} \big)$
\EndWhile
\State Compute $\hat{H_2}(p) = -2\log ({\hat{k}_t}/{\sqrt{\hat{v}_t}})$
\end{algorithmic}
\label{mainalgorithm}
\end{algorithm}

\noindent \textbf{Modeling conditional probabilities:}
Furthermore, SGM is capable of modeling conditional probabilities. We first derive the corresponding variational cost function. Let a conditional probability $p(y|x)$ and an arbitrary $p(x)$ be given. We define 
\begin{equation}
J_{Y|X} (T) = \frac{\int_\mathcal{X \times Y} T (y|x) p(y|x) p(x) d\mu d\mu}{({\int_\mathcal{X \times Y} T^2 (y|x) p(x) d\mu d\mu})^{\frac{1}{2}}}.
\label{CONDITION}
\end{equation}
Observe that its upper bound is
\begin{equation}
\begin{aligned}
J_{Y|X} (T) & = \frac{\langle T(y|x) \sqrt{p(x)},  p_\theta(y|x) \sqrt{p(x)} \rangle_{\mathcal{X} \times \mathcal{Y}}}{\langle T(y|x) \sqrt{p(x)},  T(y|x) \sqrt{p(x)} \rangle_{\mathcal{X} \times \mathcal{Y}}^{{1}/{2}}} \\
& \leq \Big({\int_\mathcal{X} \int_\mathcal{Y} p^2 (y|x) p(x) d\mu d\mu}\Big)^{\frac{1}{2}}.
\end{aligned}
\label{conditional}
\end{equation}
The RHS of~\eqref{conditional} is related to R\'enyi's definition of conditional entropy~\cite{Renyi1976some} as $H_2(Y|X) = -2 \log \sup_{T\in \mathcal{T}}J_{Y|X} (T)$. The corresponding solution is $T_0(x|y) = p(y|x)$. To use $J_{Y|X} (T)$ as a cost function for SGM, we assume that $T_0(y|x)$ is approximated by a neural network $g_\theta(y|x)$ by maximizing $J_{Y|X} (T)$. With any given $x$, the network output $g_\theta(y|x)$ defines an IMoG for the variable $y$. Let the network output be $\{w_i(\mathbf{c};x), m_i(\mathbf{c};x), A_i(\mathbf{c};x)\}$, it follows that 
\begin{equation}
\resizebox{1\linewidth}{!}{
$\begin{gathered}
\mathcal{N}_{i, j} = \mathcal{N}\big(m_i(\mathbf{c}_1;x) - m_j(\mathbf{c}_2;x), A_i(\mathbf{c}_1;x) + A_j(\mathbf{c}_2;x) \big)\\
\int_{\mathcal{Y}}g_\theta^2(y|x)p(x) d\mu = \mathbb{E}_{\mathbf{c}_1, \mathbf{c}_2, x} \big[\frac{1}{N^2}\Sigma_{i, j=1}^{N} w_i(\mathbf{c}_1;x) w_j(\mathbf{c}_2;x ) \mathcal{N}_{i, j}  \big],
\end{gathered}$}
\end{equation}
which has very little difference from the density estimation implementation. Then we maximize $J_{Y|X}(g_\theta)$ as the cost function. The details of the algorithm are shown as Algorithm~\ref{algorithmforconditional}.
\begin{algorithm}
\small
\caption{SGM for Conditional Distribution Modeling}
\begin{algorithmic}[1]
\State Initialize $k_0\leftarrow 0$; $v_0 \leftarrow 0$; $t \leftarrow 0$; Initialize $\theta_0$
\While{$\theta$ not converge}
\State $t \leftarrow t+1$
\State Sample $\{c_1, \cdots, c_{M}\}$ and $\{c_1', \cdots, c_{M}'\}$ from $U(0, 1)$ 
\State Sample $\{z_1, \cdots, z_{M}\}$ and $\{z_1', \cdots, z_{M}'\}$ from $\text{Cat}(N)$
\State Sample $\{\{x_1, y_1\}, \cdots, \{x_M, y_M\}\}$ from data distribution
\For{$i = 1, 2 \cdots M$} 
\State $w_{i} = f_{\theta_{t-1}}^{(w)}(z_i, c_i;x_i)$; $m_{i} = f_{\theta_{t-1}}^{(m)}(z_i, c_i;x_i)$; $A_{i} = f_{\theta_{t-1}}^{(A)}(z_i, c_i;x_i)$
\State $w_{i}' = f_{\theta_{t-1}}^{(w)}(z_i', c_i';x_i)$; $m_{i}' = f_{\theta_{t-1}}^{(m)}(z_i', c_i';x_i)$; $A_{i}' = f_{\theta_{t-1}}^{(A)}(z_i', c_i';x_i)$
\EndFor
\State $v_t' = \frac{1}{M}\Sigma_{i=1}^{M}  w_{i} w_{i'} \mathcal{N}(m_{i} - m_{i'}, A_{i}+A_{i'})$
\State $k_t' = \frac{1}{2M} \Big(\Sigma_{i = 1}^{M} w_{i}\mathcal{N}(m_{i} - y_{i}, A_{i}) + \Sigma_{i = 1}^{M} w_{i}'\mathcal{N}(m_{i}' - y_{i}, A_{i}') \Big)$  
\State $\hat{v}_t = \text{ADP}(v_t', t)$, $\hat{k}_t = \text{ADP}(k_t', t)$
\State $\theta_{t} = \theta_{t-1} + r \big(  \frac{\partial k_t'}{\partial {\theta_t}}/\sqrt{\hat{v}_t} -  \frac{1}{2}(\hat{k}_t \frac{\partial v_t'}{{\theta_t}} )/\sqrt{\hat{v}_t^3} \big)$
\EndWhile
\State Compute $\hat{H_2}(Y|X) = -2\log ({\hat{k}_t}/{\sqrt{\hat{v}_t}})$
\end{algorithmic}
\label{algorithmforconditional}
\end{algorithm}

\section{Experiments}

We show our experiments in two folds. First, we show how the proposed cost functions and SGM provides new opportunities to train a neural network, such as training a mutual information estimator and a GAN. Secondly, we show the powerful SGM for density estimation and how it differs and improves the conventional GMM and VBGMM.

\subsection{Training Neural Networks with Divergence}

In this section, we first show the results of using the divergence cost function~\eqref{cost2} to train neural networks, including training a mutual information estimator and a GAN.

\noindent \textbf{SGM as a mutual information (MI) estimator:} We demonstrate our methods by comparing the performance of estimating MI: (a) SGM density estimator producing the quadratic mutual information (QMI) $I_Q =-\log_2{\langle p(x, y), p(x)p(y) \rangle} + \frac{1}{2}\log_2{\langle p(x)p(y), p(x)p(y) \rangle} + \frac{1}{2}\log_2{\langle p(x, y), p(x, y) \rangle} $~\cite{principebook}; (b) Mutual information neural estimation (MINE)~\cite{belghazi2018mutual} producing Shannon's mutual information; (c) Training the same mapper as MINE with our proposed variational cost $J_{p, q}(T_\theta)$, which yields MI with R\'enyi's second-order divergence (R\'enyi's MI).

To estimate $I_Q$ with SGM, we first construct the probability ratio of the model $r_\theta(x, y) = \frac{g_\theta(x, y)}{g_\theta(x)g_\theta(y)}$. We define a function of the probability ratio that evaluates MI of the model given data as $\hat{I}_Q(r_\theta) = -\frac{1}{2}\log_2 \bigl( \frac{\mathbb{E}_{P_X P_Y} [r_\theta(x, y)]}{\mathbb{E}_{P_{XY}} [r_\theta(x, y)] }\bigr)$. It can be verified $\hat{I}(r_\theta)$ is an unbiased estimator of $I_Q$ as $r_\theta \rightarrow \frac{p(x, y)}{p(x)p(y)}$. Each term in $\hat{I}_Q(r_\theta)$ is a byproduct from optimization. 

We generate a mixture distribution with 20 centers uniformly sampled from 0.2 to 0.8 in 2D space. We assign each center to a Gaussian distribution with diagonal convariance matrices from $U(0.0001, 0.002)$. We sample the weighting factor from $U(0.5, 1.5)$. We set $N=300$ across the experiments.

We test three approaches for the MoG example as before and estimate the mutual information between two dimensions. Figure~\ref{MINE1} shows both our approaches are far more stable than MINE which will diverge for small number of samples. Even with variance reduction, MINE suffers from a large variance that cannot be ignored. We compute the ground truth of QMI through~\eqref{qudratic_form} and SGM matches the exact value with $5$k samples. R\'enyi's MI also converges at the same rate, but does not converge to the same value because it uses R\'enyi's divergence, instead of QMI, which is a Cauchy-Schwarz divergence~\cite{jenssen2006cauchy} estimated with R\'enyi's second-order entropy. This also shows that the proposed cost can be applied to any neural network. MINE with enough samples also converges to Shannon's mutual information but with a high variance. So these proposed estimators are much more well behaved and data efficient to estimate MI, outperforming MINE. 

To show the scalability to high dimensions, we generate MoG in high dimensions and compare MINE with optimizing $J_{p,q}(T_\theta)$, shown as~Figure~\ref{MINE2}. This shows our new cost has the same performance with less variance, but much more stable when only a limited number of samples are accessible, as demonstrated in Figure~\ref{MINE1}.

\noindent\begin{minipage}{1\linewidth}
\centering
\captionsetup{type=figure}
\subfloat[{\small Sample Efficiency of MI Estimators}]{\includegraphics[width=0.9\linewidth]{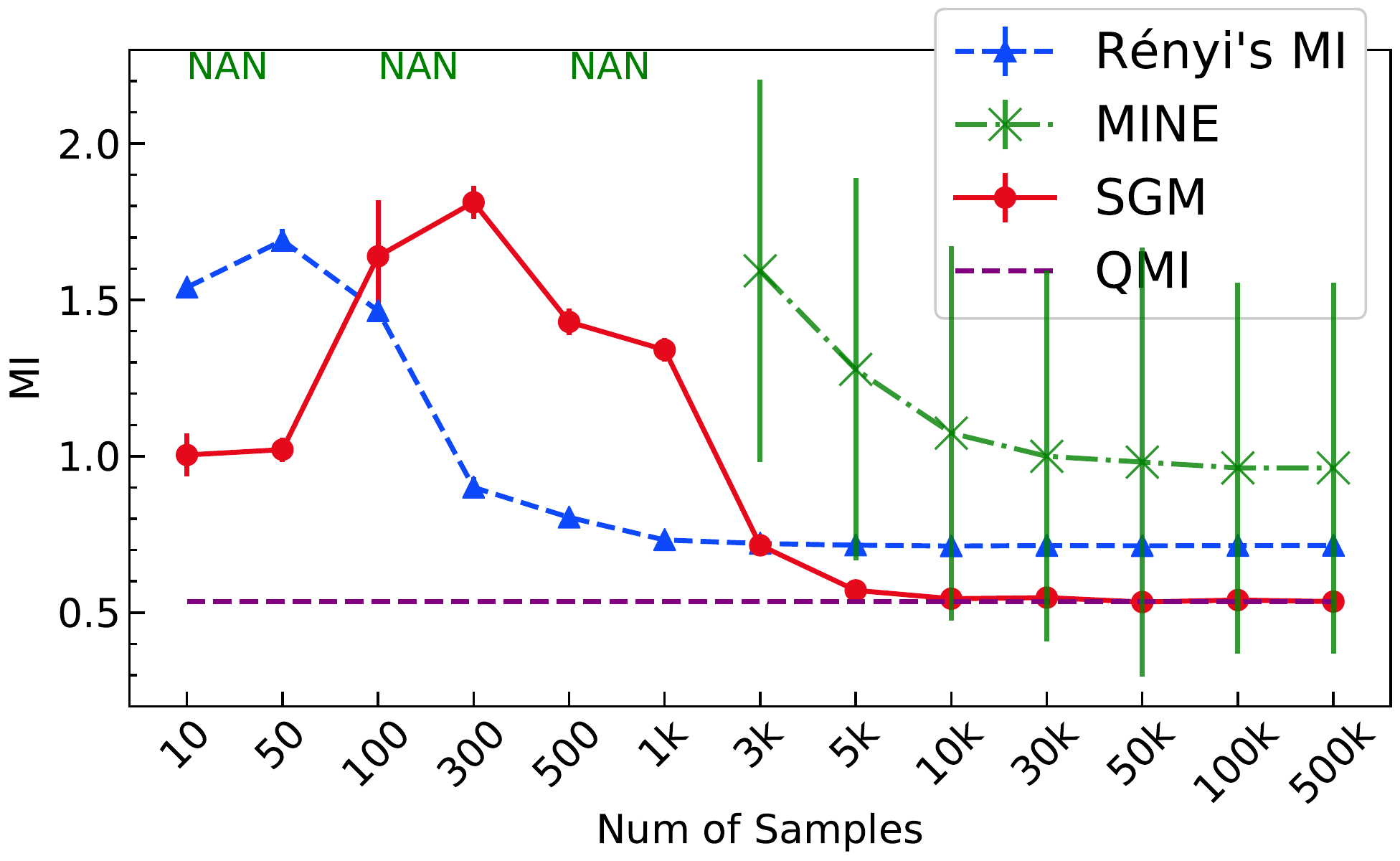}\label{MINE1}}\\
\subfloat[{\small Scalability of MI Estimators}]{\includegraphics[width=0.9\linewidth]{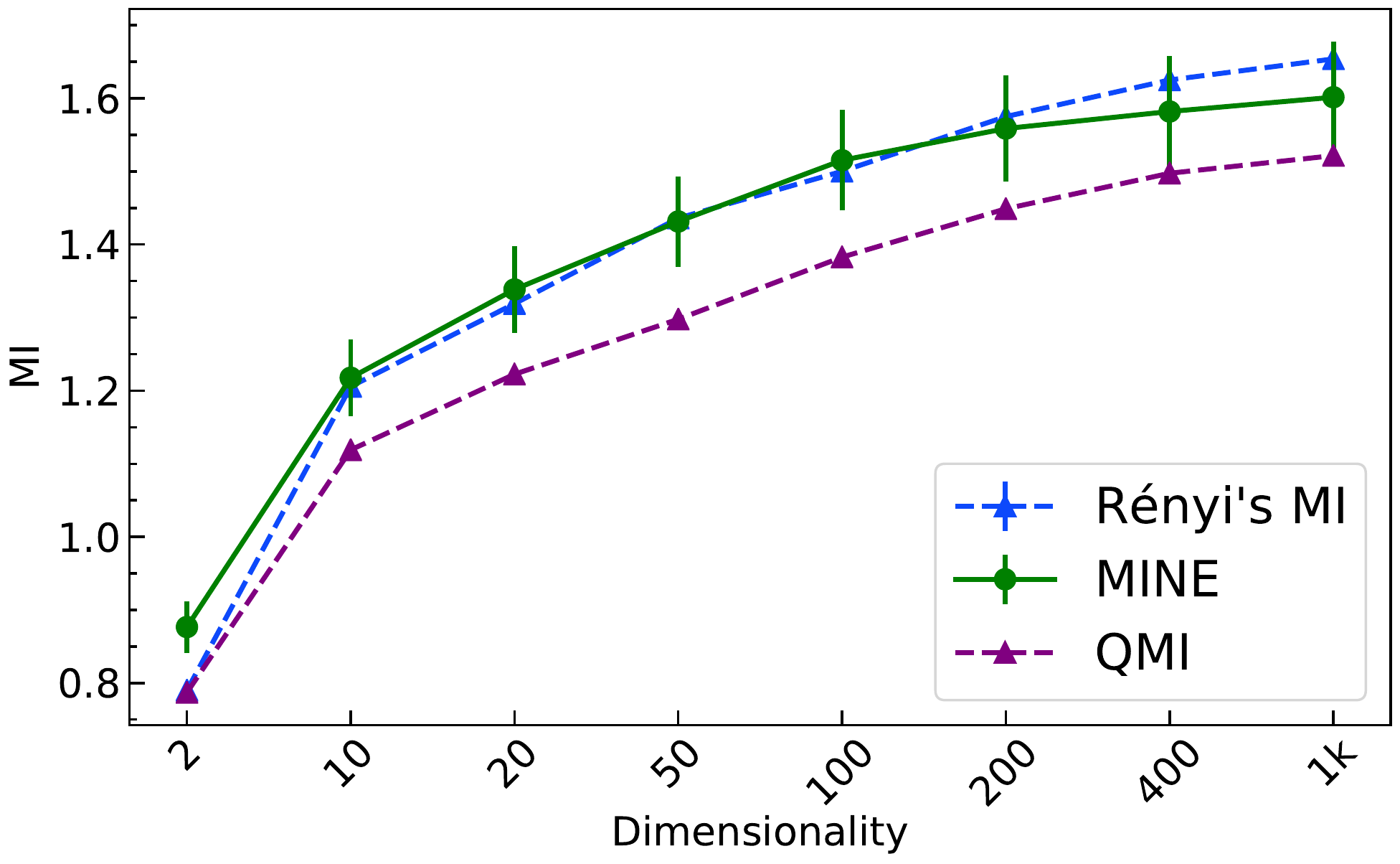}\label{MINE2}}
\captionof{figure}{{\small Performance of MI estimators. (a) shows both SGM and optimizing $J_{p,q}(T_\theta)$ shows a better stability and lower variance than MINE, where MINE will diverge (producing ``NAN'') for small number of samples. Three approaches converge to R\'enyi's second-order MI, Shannon's MI and QMI respectively. (b) shows that optimizing $J_{p, q}(T_\theta)$ also scales up to high dimensions with a lower variance. As we decrease the training samples for MINE, it will no longer converge.}}
\end{minipage}\\

\noindent \textbf{A new approach to train a GAN:} We show the possibility of using $J_{p, q}(T)$ to train a GAN. As stated, the corresponding solution of optimizing a variational form of divergence yields a function of probability ratio, which is rarely addressed in the GAN literature. Let $T_\theta:\mathcal{X} \rightarrow \mathbb{R}$ be the discriminator network and $g_\theta$ be the generator network, we formulate the optimization problem as
\begin{equation}
\text{min}_{g_\theta} \; \text{max}_{T_\theta} \;\; J_{ g_\theta, p}(T_\theta) = \mathbb{E}_{g_\theta}[T_\theta(x)]/({\mathbb{E}_p[T_\theta^2(x)]})^{\frac{1}{2}}.
\label{gan_optimum}
\end{equation}
For each fixed $g_\theta$, the corresponding optimal solution of $T_\theta$ is $T_0(x) = \beta {g_\theta(x)}/{p(x)}$. Ideally, we should obtain $T_0(x) = 0$ for any $x \sim p$. 

We train a conventional GAN on the Fashion MNIST. The generated samples are shown as Figure~\ref{generated}. Figure~\ref{fpimage} shows the false positive and false negatives with a threshold of $10^{-2}$, which shows the exemplars closer to the boundary. Figure~\ref{indis} shows the discriminator network output for the Fashion MNIST test set. Then we apply the same discriminator network to the test set of MNIST, producing Figure~\ref{ood}, where the network produces 0 for in-distribution samples and 1 for out-of-distribution (OOD) samples. We compare our approach with baselines mentioned in~\cite{ren2019likelihood}, including maximum class probability~\cite{hendrycks2016baseline}, entropy, Mahalanobis distance~\cite{lee2018simple}, likelihood ratio~\cite{ren2019likelihood}. Additionally, we also compare with two state-of-the-art methods based on variational information bottleneck (VIB)~\cite{alemi2018uncertainty} and nonlinear information bottleneck (NIB)~\cite{kolchinsky2019nonlinear}, respectively. In terms of metrics, we use the area under ROC curve (AUROC$\uparrow$), the area under precision-recall curve (AUPRC$\uparrow$), and the false positive rate at
$80\%$ true positive rate (FPR$80\downarrow$). The quantitative results are summarized in Table~\ref{table_OOD_detection}.

\noindent\begin{minipage}{1\linewidth}
\begin{centering}
\captionsetup{type=figure}
\centering
\subfloat[{\small Generated}]{\includegraphics[width=0.3\textwidth]{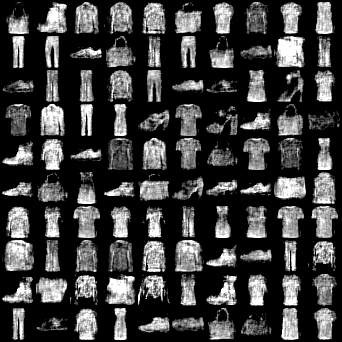}\label{generated}} \hspace{10pt}
\subfloat[{\small FP/FN Samples}]{\includegraphics[width=0.3\textwidth]{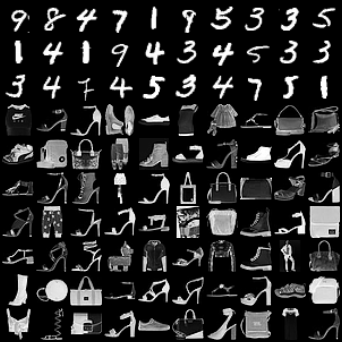}\label{fpimage}} \\[-10pt]
\subfloat[{\small In-Distribution Ratio}]{\includegraphics[width=0.4\textwidth]{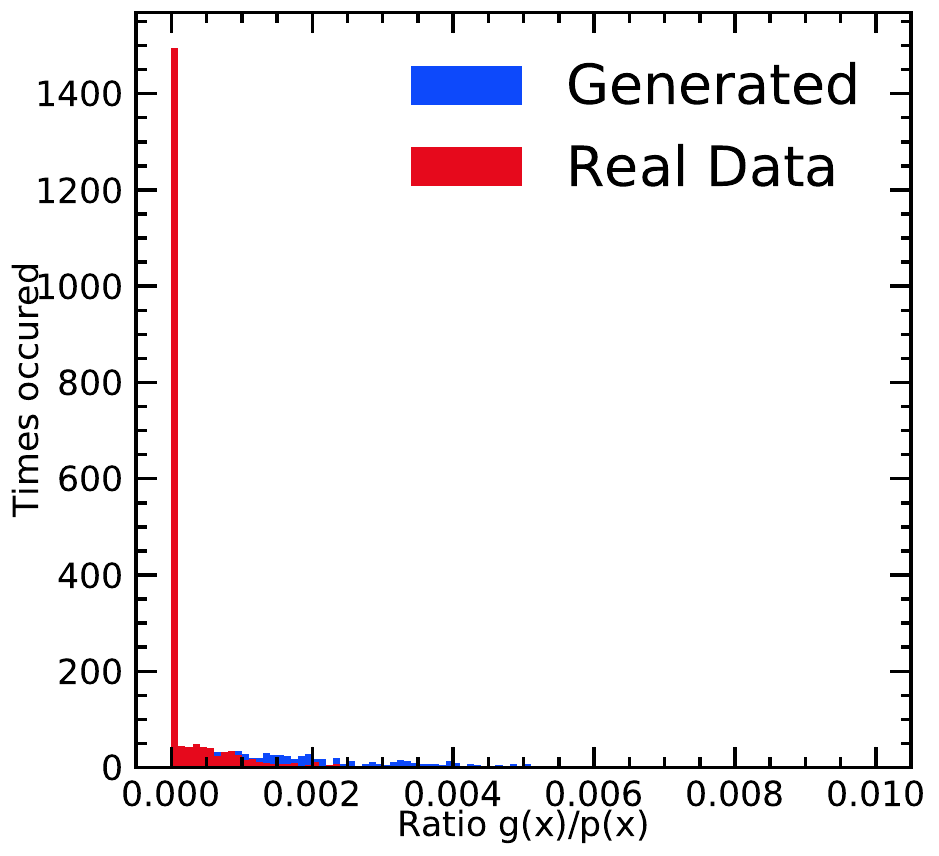}\label{indis}}
\subfloat[{\small OOD Ratio}]{\includegraphics[width=0.4\textwidth]{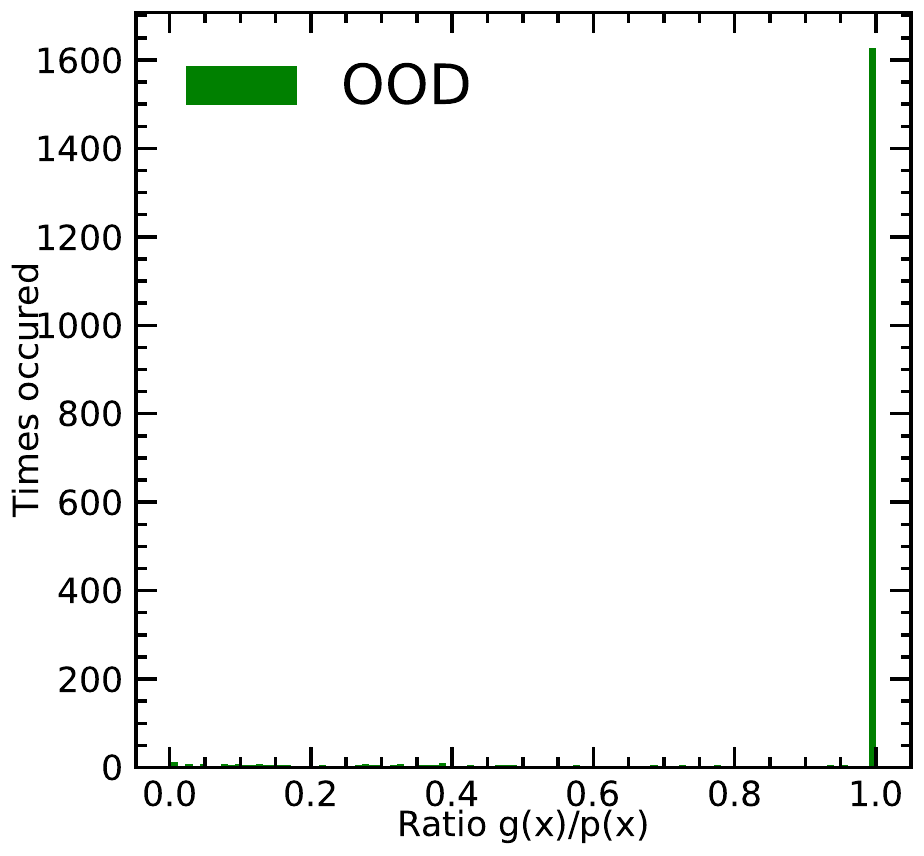}\label{ood}}
\vspace{-5pt}
\captionof{figure}{\small Network outputs trained by~\protect\eqref{gan_optimum}. (a) shows the generated samples. (b) shows the FP and FN samples when the ratio is set to $10^{-2}$. (c) and (c) show the produced probability ratio by the discriminator network for generated data, Fashion MNIST test set, and MNIST test set. }
\end{centering}
\end{minipage}\\

\noindent \begin{minipage}{1\linewidth}
\centering
\captionof{table}{\small AUROC$\uparrow$, AUPRC$\uparrow$, and FPR ($80\%$TPR) $\downarrow$ for detecting OOD inputs with density ratio and other baselines\protect\footnote{Results of $p(\hat{y}|\mathbf{x})$, entropy of $p(y|\mathbf{x})$, Mahalanobis distance and likelihood ratio are from~\protect\cite{ren2019likelihood}. Results of VIB and NIB are obtained on our test environment with authors’ original code.} on Fashion-MNIST vs. MNIST datasets. $\uparrow$ indicates that larger value is better, $\downarrow$ indicates that lower value is better.}
\resizebox{0.9\linewidth}{!}{
\large
\begin{tabular*}{300pt}{@{\extracolsep{\fill}} c ccc}
\toprule[2pt]
     Method & AUROC$\uparrow$ & AUPRC$\uparrow$ & FPR$80$ $\downarrow$ \\
\midrule
      Density ratio (ours) & $\textbf{0.997}$ & $\textbf{0.997}$ & $0.004$ \\
    $p(\hat{y}|\mathbf{x})$ & $0.734$ & $0.702$ & $0.506$ \\
   Entropy of $p(y|\mathbf{x})$ & $0.746$ & $0.726$ & $0.448$  \\
   Mahalanobis distance & $0.942$ & $0.928$ & $0.088$ \\
   Likelihood ratio & $0.994$ & $0.993$ & $\textbf{0.001}$ \\
   VIB & $0.906$ & $0.903$ & $0.172$ \\ 
   NIB & $0.916$ & $0.913$ & $0.152$ \\
\bottomrule[2pt]
\end{tabular*}
\label{table_OOD_detection}}
\end{minipage}

\subsection{Density Estimation}
Now we show how density estimation performance of SGM. We extend the MoG example for mutual information estimations to a generalized mixtures. We randomly assign each center with a Gaussian distribution, a Laplacian distribution, or a uniform distribution. We sample the scale of the Laplacian from $U(0.01, 0.2)$, the subinterval length of each uniform distribution from $U(0.01, 0.2)$, and the weighting factor from $U(0.5, 1.5)$. We generate 200k samples in total and repeat the experiment with different parameters for five times. We compare SGM with GMM and VBGMM. VBGMM has the concentration factor 1. \\

\noindent \textbf{Evaluations:} We use $J_p(g_\theta)$ (CE) and the validation rate for evaluation. We compute CE with~\eqref{discrete} for GMM, with~\eqref{infinite_gauss_form} for SGM and VBGMM. A greater CE indicates a better performance. To address the overfitting issue of GMM, suppose $m'(j)$ is the $j$-th true center, we use the validation rate $\text{VR} = \mathbb{E}[\mathbf{w} \cdot \mathds{1}((\min_{
j}\|\mathbf{m}(x_t) - m'(j)\|^2_2) <d) ]/\mathbb{E}[\mathbf{w}]$ to calculate the percentage of the components whose mean's distance to the closet true center mass is under $d = 10^{-4}$ or~$10^{-2}$.\\

\noindent\textbf{Performance:} Table~\ref{tablemog} shows that the performance of SGM is higher than both GMM and VBGMM. If we estimate CE's upper bound~\eqref{global} with the true pdf by subintervals, we obtain $6.68$, $21.0$, $12.5$, $7.26$ and $8.25$. Since only finite samples are given, the true values are higher. Although VBGMM has a better concentration, it introduces a huge bias regarding performance. With $d$ chosen to be $10^{-2}$, all the means of SGM are valid while GMM suffers from severe overfitting, illustrated as Figure~\ref{figmog}.

\noindent\begin{minipage}{1\linewidth}
\captionsetup{type=figure}
\centering
\subfloat[]{\includegraphics[width=0.25\textwidth]{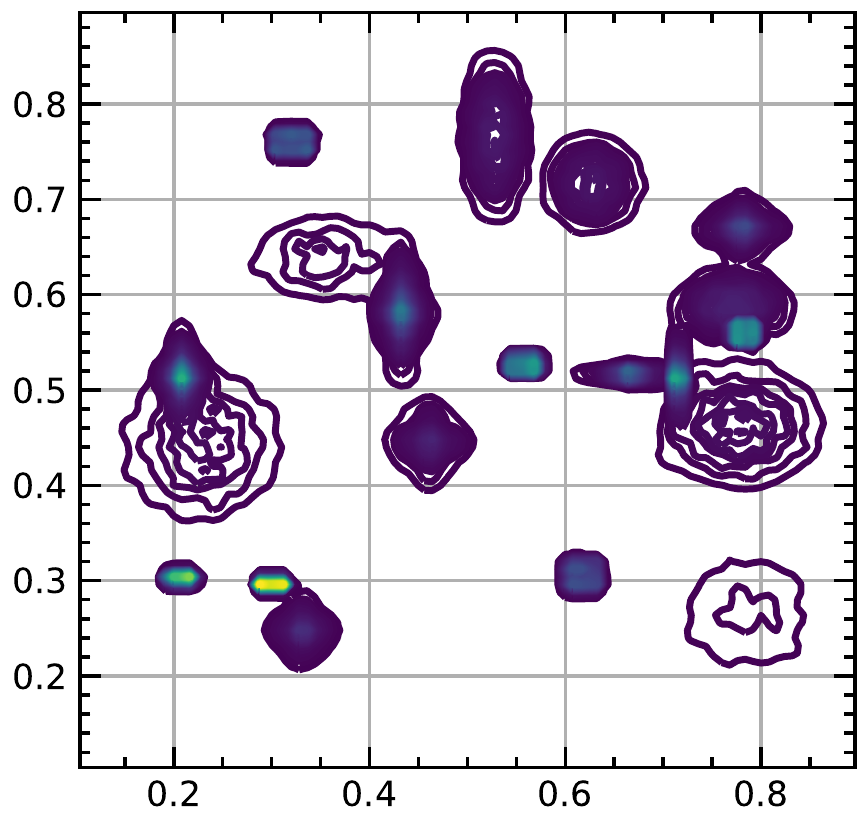}}
\subfloat[]{\includegraphics[width=0.25\textwidth]{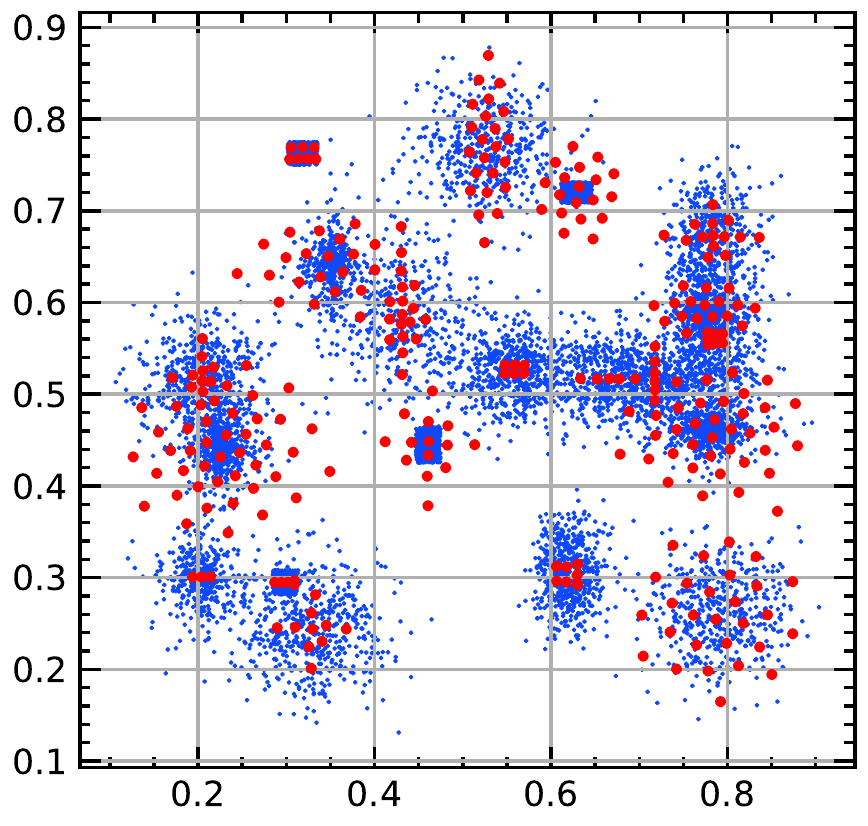}}
\subfloat[]{\includegraphics[width=0.25\textwidth]{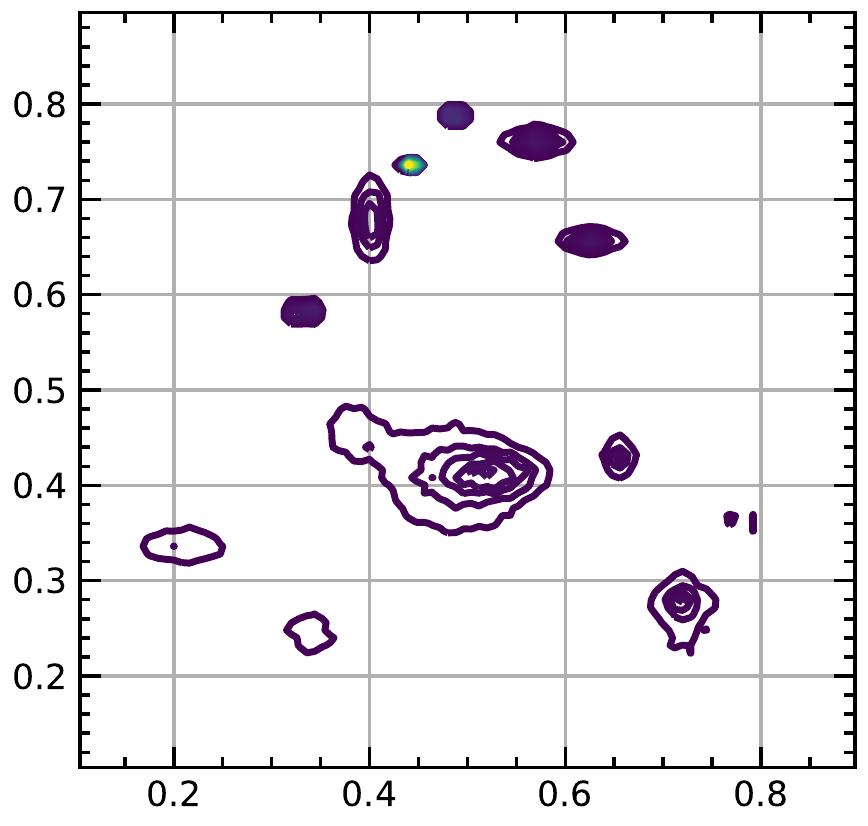}}
\subfloat[]{\includegraphics[width=0.25\textwidth]{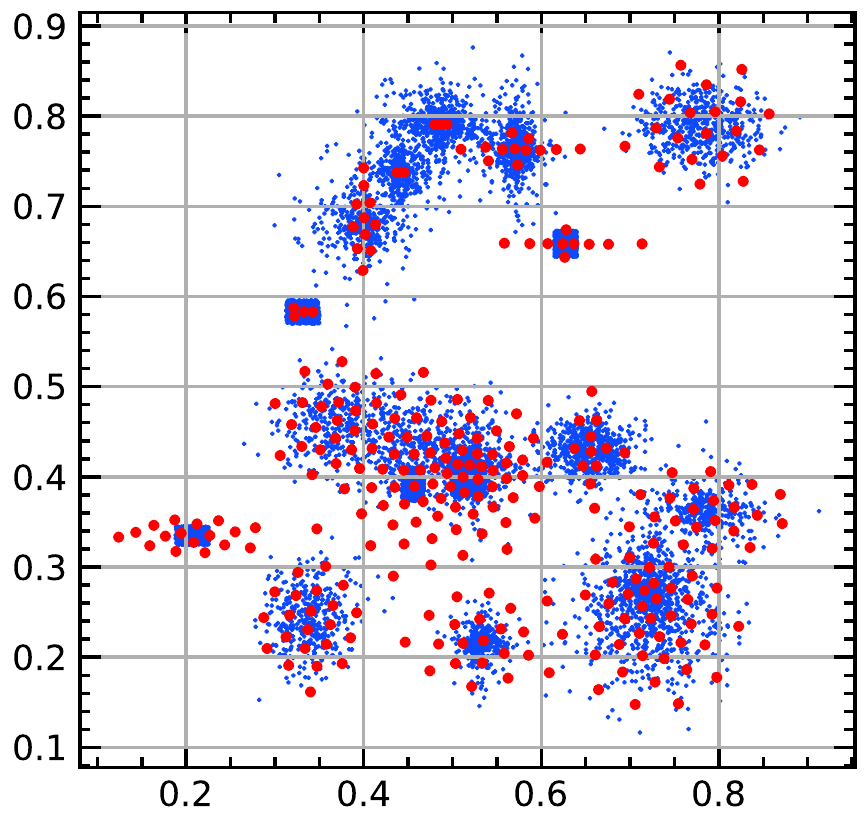}}\\[-10pt]
\subfloat[]{\includegraphics[width=0.25\textwidth]{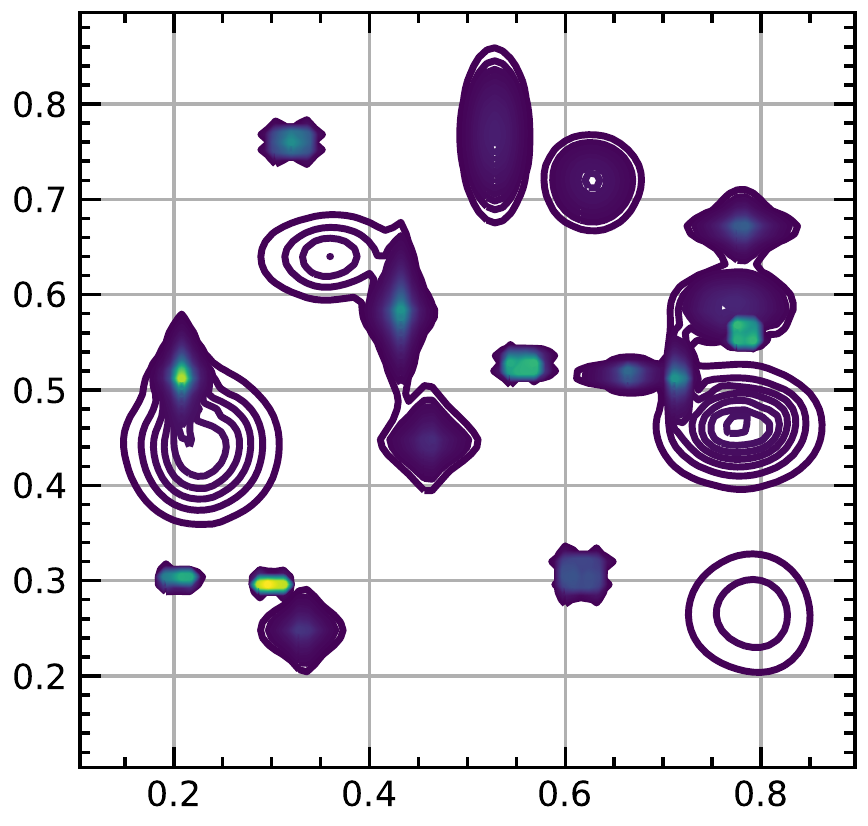}}
\subfloat[]{\includegraphics[width=0.25\textwidth]{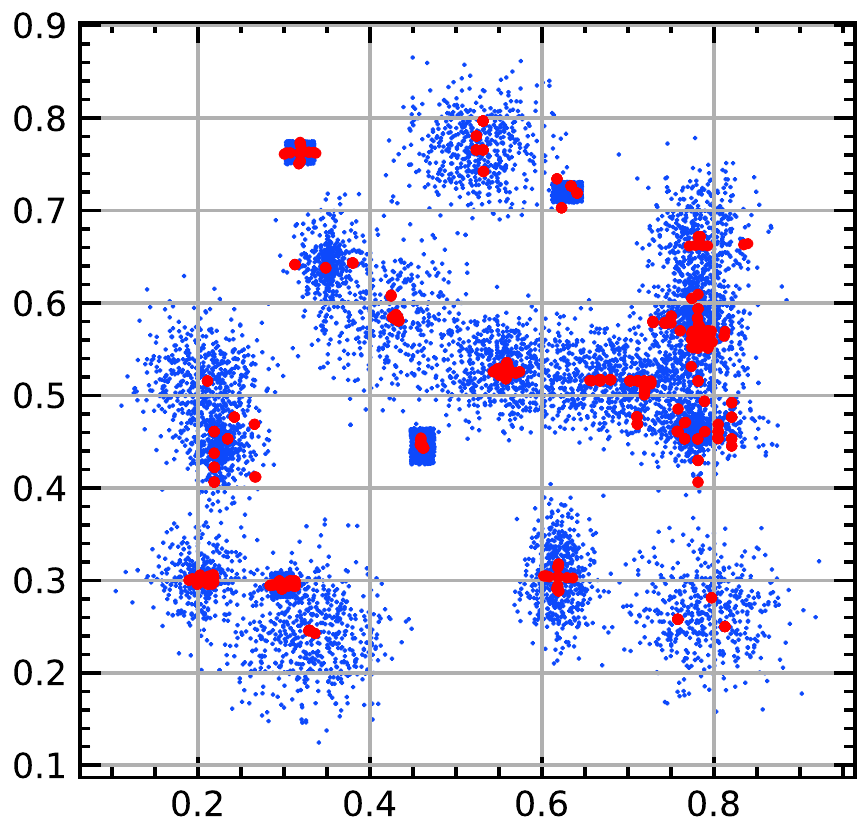}}
\subfloat[]{\includegraphics[width=0.25\textwidth]{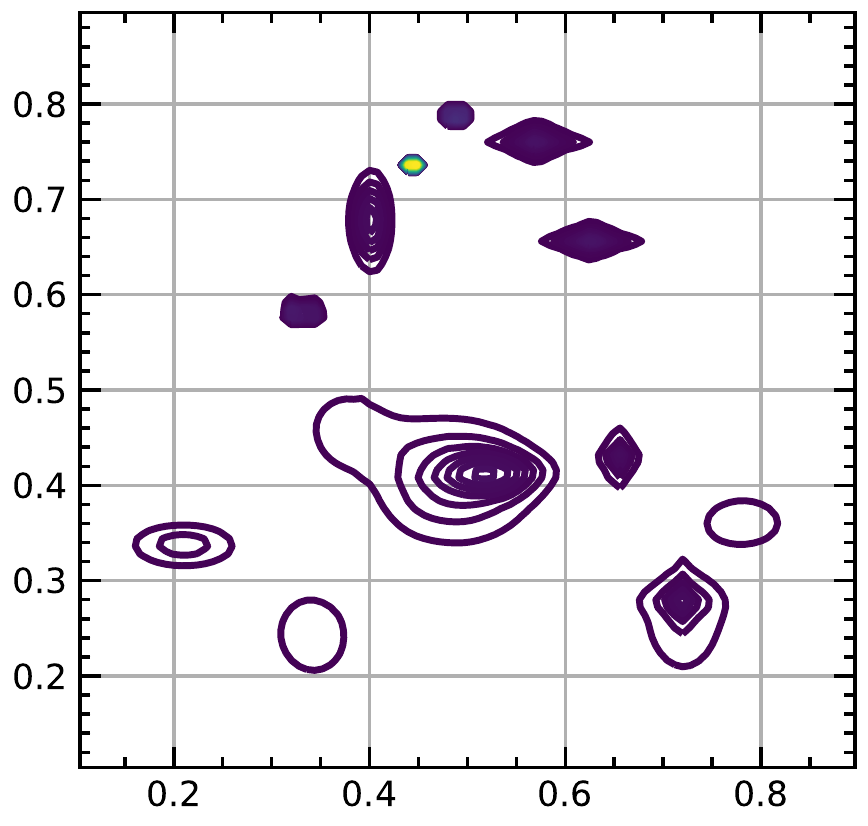}}
\subfloat[]{\includegraphics[width=0.25\textwidth]{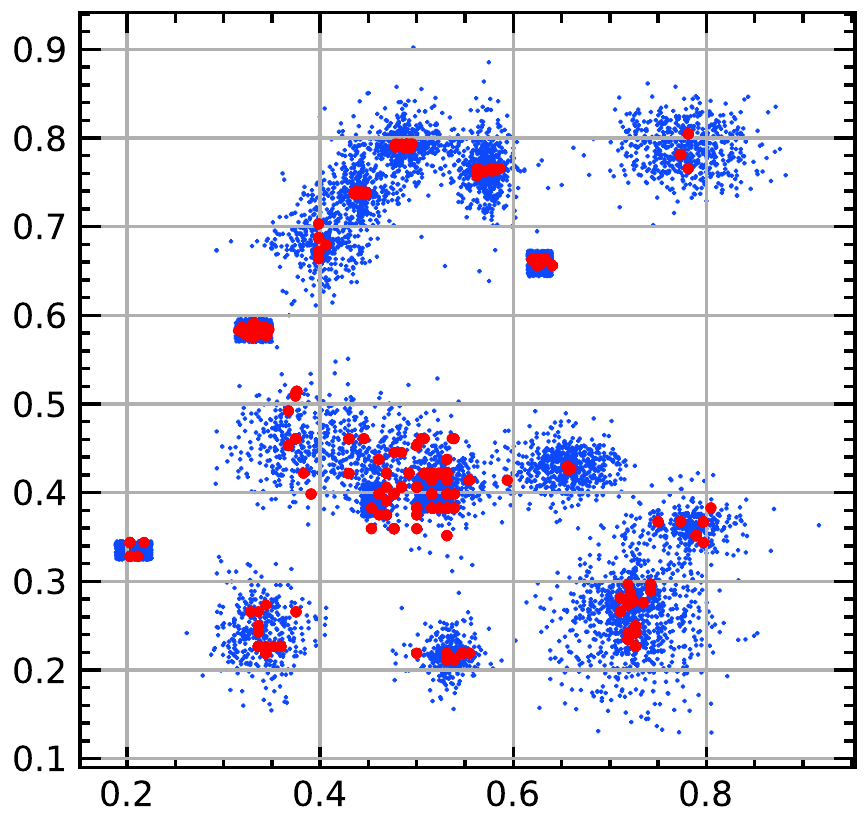}}
\vspace{-8pt}
\captionof{figure}{\small (a) to (d) show the density and model mean components learned by GMM. (e) to (h) show the same figures produced by SGM. The red dots represent the model means and the blue dots represent the data. SGM has a better performance regarding tail.}
\label{figmog}
\end{minipage}

\noindent\begin{minipage}{1\linewidth}
\large
\centering
\captionof{table}{{\small Performance of the 2D density estimation. Although VBGMM has a better concentration, it suffers from a much higher bias compared to vanilla GMM and SGM.}}
\resizebox{0.8\linewidth}{!}{
\begin{tabular*}{320pt}{@{\extracolsep{\fill}} c ccccc}
\toprule[3pt]
     \tnote{a}  Algorithm\tnote{b} & 
     \multicolumn{5}{c}{CE}  \\  \cmidrule{2-6} & EXP \#1 & EXP \#2 & EXP \#3 & EXP \#4 & EXP \#5  \\
\midrule
      GMM&
      6.40&17.31&13.36&7.14&8.06  \\
      VBGMM&
      4.13 &4.86 &4.33 &4.19 &5.48 \\
      SGM& \textbf{6.64} & \textbf{18.02} & \textbf{13.80} &\textbf{7.76} & \textbf{9.69} \\
 \midrule[3pt]
      \tnote{a}  \tnote{b} & 
     \multicolumn{5}{c}{VR $d=10^{-4}$}  \\  \cmidrule{2-6} & EXP \#1 & EXP \#2 & EXP \#3 & EXP \#4 & EXP \#5  \\
\midrule
      GMM&
      0.0036&0.018&0.012&0.015&0.23  \\
      VBGMM&
      \textbf{0.74} &\textbf{0.84} &\textbf{0.84} &\textbf{0.71} &\textbf{0.89} \\
      SGM& 0.57 &0.24 &0.40 &0.36 &0.37 \\
 \midrule[3pt]
     \tnote{a}  \tnote{b} & 
     \multicolumn{5}{c}{VR $d=10^{-2}$}  \\  \cmidrule{2-6} & EXP \#1 & EXP \#2 & EXP \#3 & EXP \#4 & EXP \#5  \\
\midrule
      GMM&
      0.86&0.91&0.87&0.61&\textbf{1.0}  \\
      VBGMM&
      \textbf{1.0}&\textbf{1.0}&\textbf{1.0}&\textbf{1.0}&\textbf{1.0}\\
      SGM& \textbf{1.0}&\textbf{1.0}&\textbf{1.0}&\textbf{1.0}&\textbf{1.0}\\
       \bottomrule[3pt]
\end{tabular*}
}\label{tablemog}
\end{minipage}

\subsection{Modeling Conditional Probabilities}
Next we show that SGM is capable of modeling conditional probabilities. We assume that the conditional is given by an artificial continuous-state-space Markov chain. We create a $10\times 10$ matrix with main diagonals to be $0.7$ and the other entries to be $1/30$. Then we shuffle the matrix along the $x$ axis and put Gaussian distributions on top of each center, with diagonal covariance matrices sampled from $U(0.0005, 0.002)$ to formulate the single transition probability as $p(x_{t+1}|x_t) = {p(x_t, x_{t+1})}/{\int_\mathcal{X} p(x_t) d\mu}$. We start with sampling 100 points from a uniform distribution between 0 and 1. Then we generate trajectories with a length of 1000. In other words, samples will have a joint distribution $p(x, y) = p(x)p(x_{t+1}=y|x_t=x)$ with the marginal $p(x) = \mathbb{E}_{x_0}[\frac{1}{T}  \sum_{t=0}^T p(x_t = x)]$.\\

\noindent\textbf{Evaluations:} We compare SGM with GMM, VBGMM, MDN and conditional GAN (CGAN). For GMM and VBGMM, we estimate the joint $p(x, y)$ as described and then marginalize $x$ to obtain $p(y|x) = p(x, y) / \int_X p(x, y) d\mu$. For MDN we use the implementation in~\cite{bishopmixture}. We also implement CGAN~\cite{mirza2014conditional}. Given two pairs $\{x_t, x_{t+1}\}$ and $\{x_t', x_{t+1}'\}$, the discriminator tries to distinguish between $\{x_t, g_\theta(x_t)\}$ and $\{x_t', x_{t+1}'\}$. SGM is trained by~\eqref{CONDITION}. For GMM and VBGMM, suppose the covariance matrices are diagonal, notice that $g_\theta(y|x) = \frac{\sum_{j=1}^Nw_j\mathcal{N}(x-\mu_{x_j}, \sigma_{x_j})\mathcal{N}(y-\mu_{y_j}, \sigma_{y_j})}{\sum_{i=1}^N w_i\mathcal{N}(x-\mu_{x_i}, \sigma_{x_i})}$. So we first compute the responsibility for each $x\sim p$, then we compute $J_{Y|X}(g_\theta)$ with~\eqref{CONDITION}.\\

\noindent\textbf{Performance:} Figure~\ref{figurerate} shows the performance of each method. We find that similar to GMM, MDN also has the tendency to overestimate tail. Since the original joint distribution is a MoG, VBGMM is the most competitive but still has visible bias. For CGAN, we use the empirical estimator to produce the joint distribution as Figure~\ref{cganj} and the conditional as Figure~\ref{cganc}. Since CGAN does not assume the shape of the distribution and its input noise is uniformly distributed, the estimation will always have a bias. Table~\ref{tablerate} shows the overall score evaluated with CE. 

\section{Conclusion}
This paper presents a novel framework for generative modeling, named SGM, exploiting the nice properties of R\'enyi’s information in estimation. Since mixtures of Gaussians can be estimated in closed-form by R\'enyi’s quadratic information, we derive two new cost functions that substitute the cross-entropy and the KL divergence by the corresponding R\'enyi’s formulations. We then proceed to train a neural network that directly yields a generative model of the input data by local approximations of mean and variance in the data space, e.g., under an infinite Gaussian mixture model assumption.  This avoids the issue of using bounds (upper or lower) in the conventional generative modeling approaches, which requires assumptions that are very difficult to fulfill in parametric density modeling. The proposed SGM is data-efficient unlike the conventional approaches as we clearly show with the comparison with MINE. In the experiments, we verified the accuracy of the proposed SGM approach. Further work will be directed towards finding alternate approximations to $T(x)$, which will scale better for high dimensions, and extend the method to alpha close to 1 to estimate Shannon information. 

\noindent\begin{minipage}{1\linewidth}
\captionsetup{type=figure}
\centering
\subfloat[GT]{\includegraphics[width=0.25\textwidth]{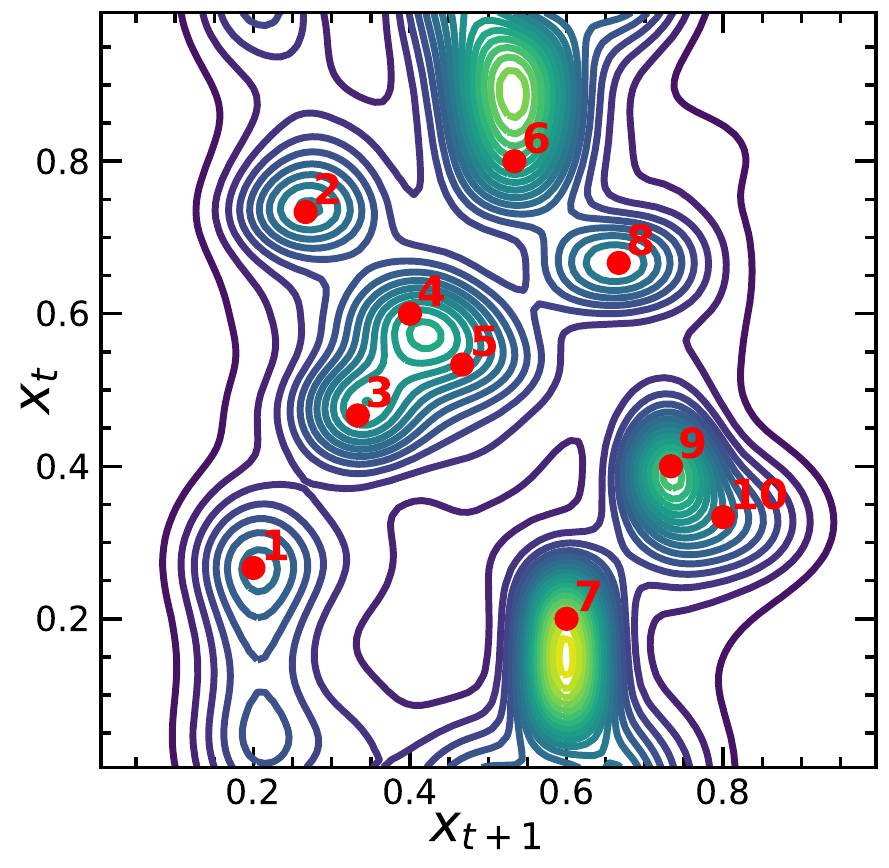}}
\subfloat[SGM]{\includegraphics[width=0.25\textwidth]{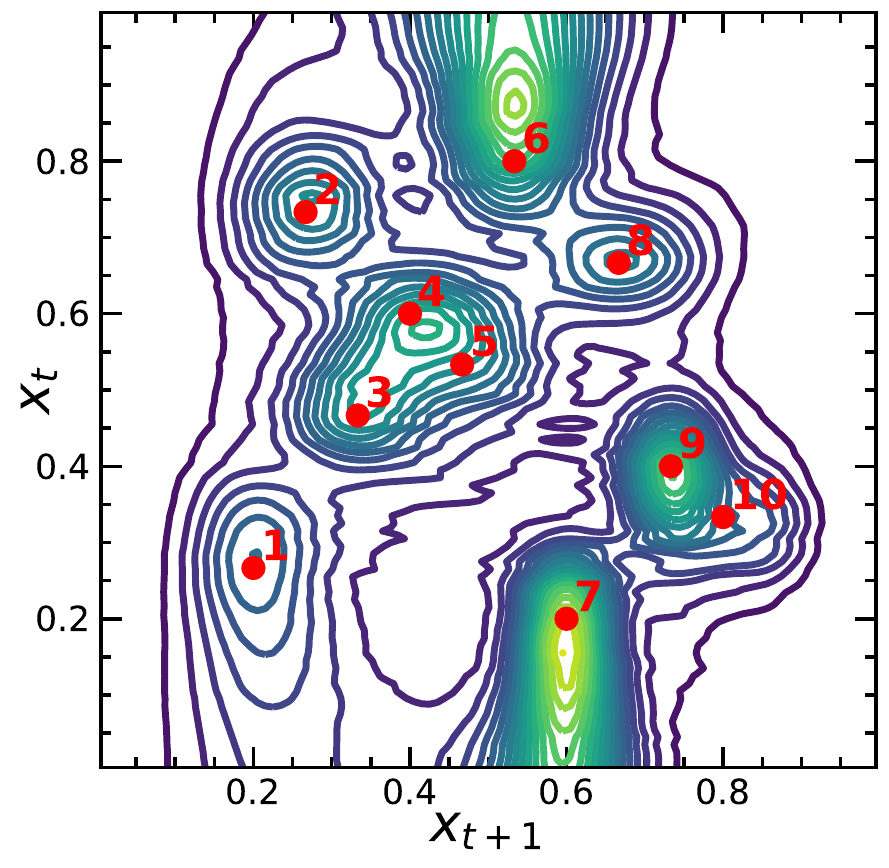}}
\subfloat[GMM]{\includegraphics[width=0.25\textwidth]{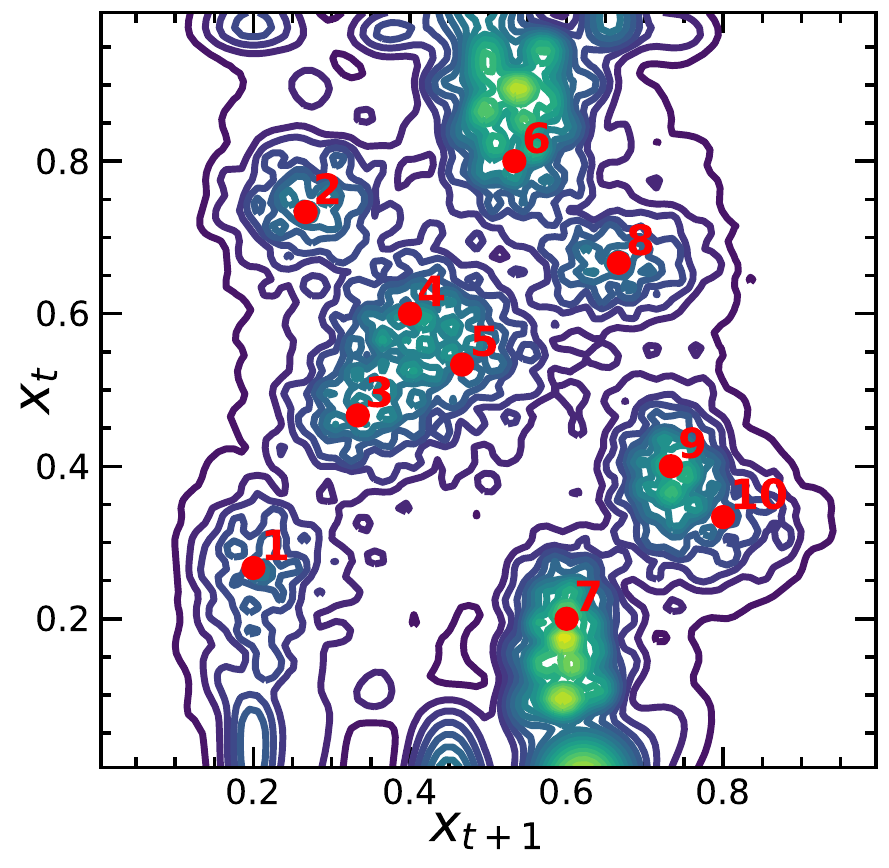}}
\subfloat[VBGMM]{\includegraphics[width=0.25\textwidth]{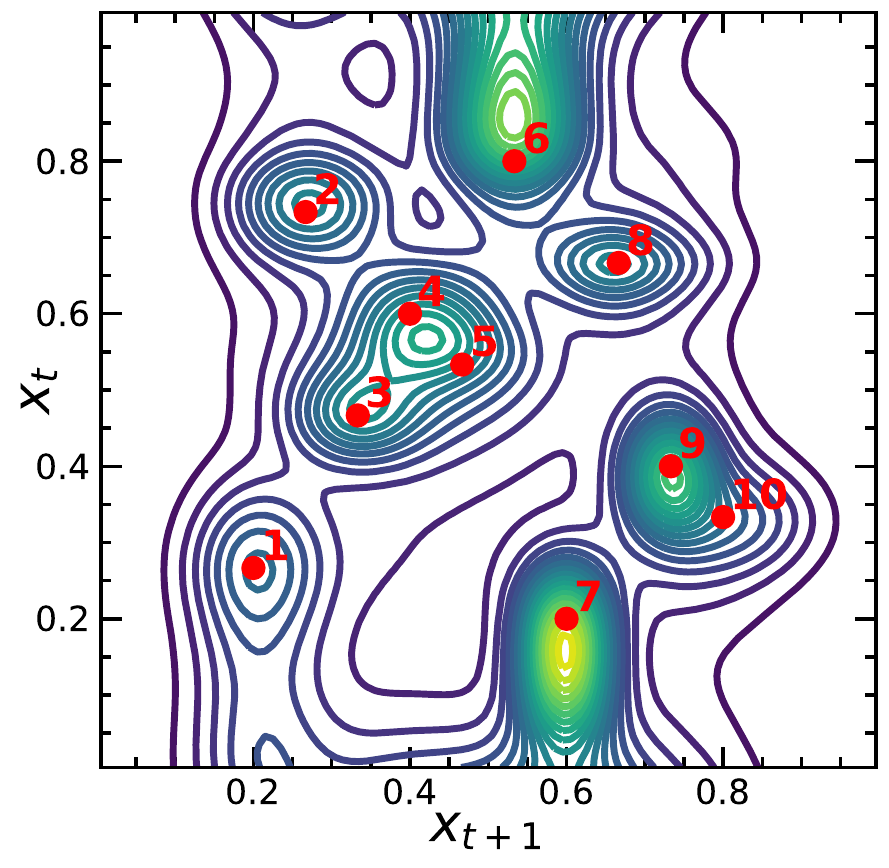}}\\[-10pt]
\subfloat[MDN]{\includegraphics[width=0.25\textwidth]{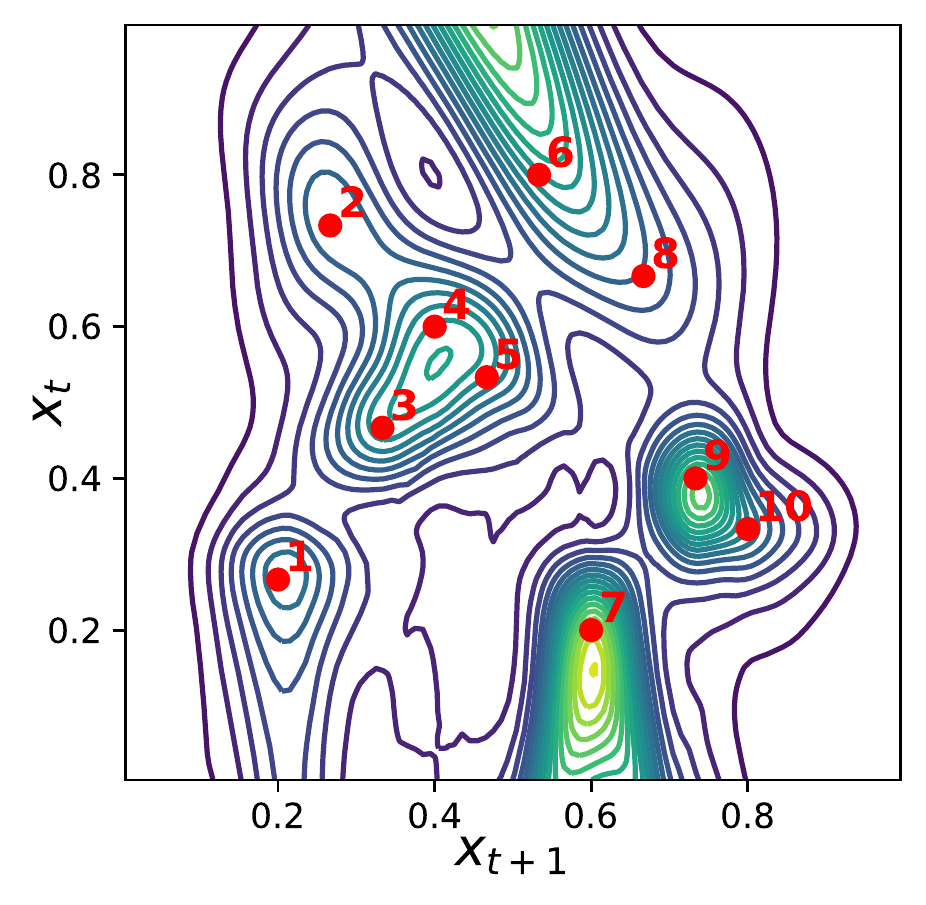}}
 \subfloat[Plugged-in]{\includegraphics[width=0.25\linewidth]{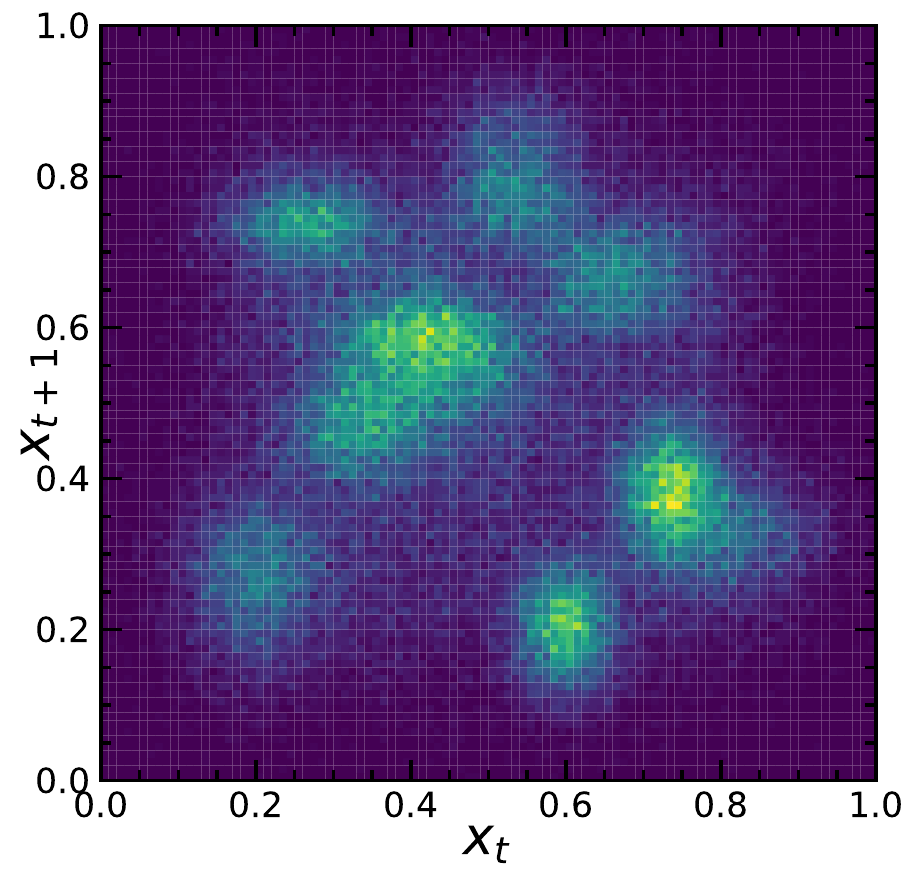}\label{dataaa}}
\subfloat[CGAN-j]{\includegraphics[width=0.25\linewidth]{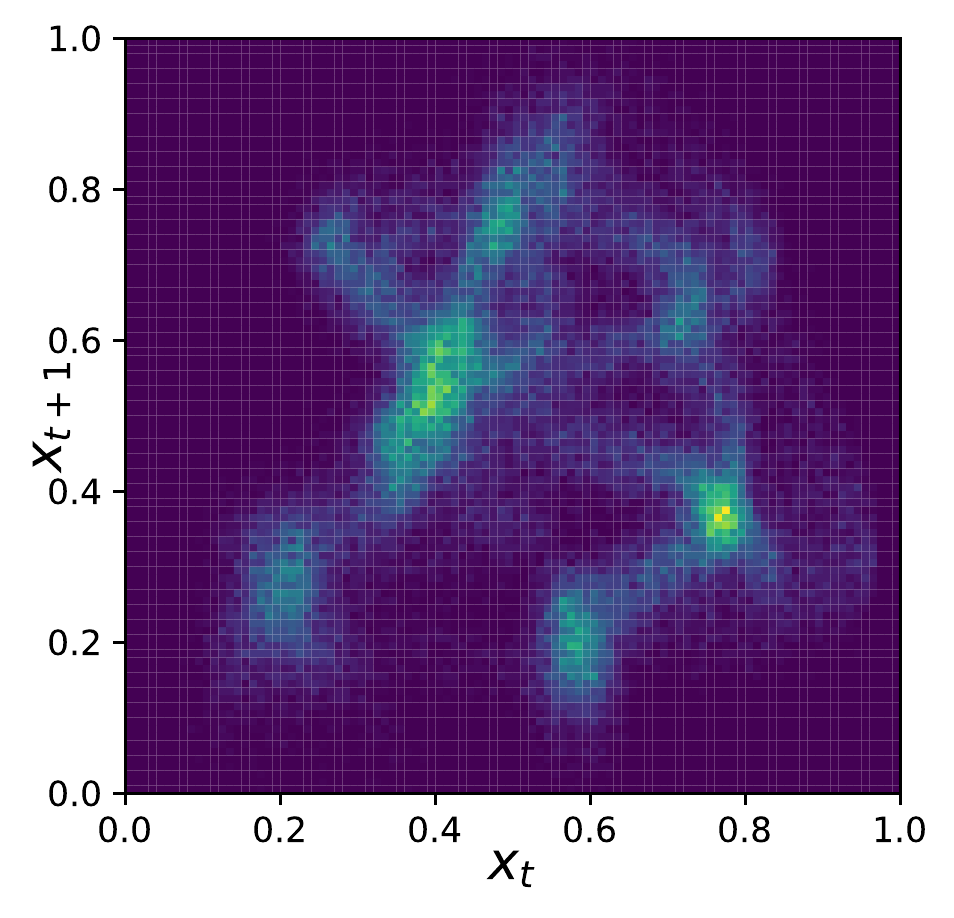}\label{cganj}}
\subfloat[CGAN-c]{\includegraphics[width=0.25\linewidth]{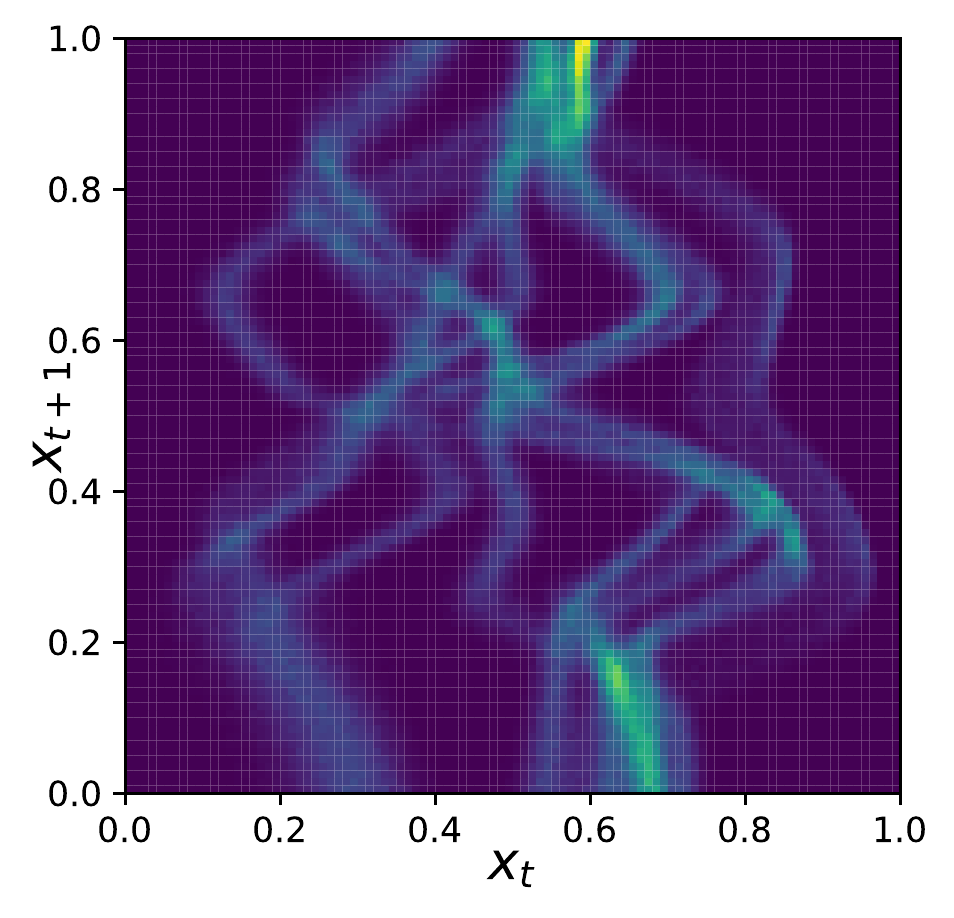}\label{cganc}}
\captionof{figure}{\small Comparisons of learned transition probabilities. The red dots show the transition with the highest probability at each state when constructing the Markov chain. (g) shows the joint distribution learned by CGAN. (h) shows $p(x_{t+1}|p_t)$ by taking the network input to be evenly spaced between 0 and 1.}
\label{figurerate}
 \end{minipage}

\noindent\begin{minipage}{1\linewidth}
\large
\centering
\captionof{table}{\small Performance for conditional probability modeling.}
\resizebox{0.9\textwidth}{!}{
\begin{tabular*}{300pt}{@{\extracolsep{\fill}} c ccccc}
\toprule[3pt]
     \tnote{a}  Algorithm\tnote{b} & 
     \multicolumn{5}{c}{CE}  \\  \cmidrule{2-6} & EXP \#1 & EXP \#2 & EXP \#3 & EXP \#4 & EXP \#5  \\
\midrule
      CGAN&
      0.788 &0.796&0.675  &0.659 &0.887  \\
      GMM&
      0.900&0.939&0.878 &0.957 &0.980  \\
      VBGMM&
      0.959 &0.942 &0.904  &0.945  &0.951  \\
      MDN&
      0.967 &0.957 &0.889  &0.913  &0.912  \\
      SGM& \textbf{0.993} &\textbf{0.968} &\textbf{0.923}  &\textbf{0.972}  &\textbf{0.990} \\ 
      \bottomrule[3pt]
\end{tabular*}
\label{tablerate}}
\end{minipage}%

\bibliography{references}
\bibliographystyle{unsrt}

\end{document}